\documentclass[10pt,twocolumn,letterpaper]{article}

\usepackage{cvpr}              

\definecolor{cvprblue}{rgb}{0.21,0.49,0.74}
\usepackage[pagebackref,breaklinks,colorlinks,allcolors=cvprblue]{hyperref}

\usepackage{hyperref}
\usepackage{url}
\usepackage{wrapfig}
\usepackage{graphicx} 
\usepackage{algorithm}
\usepackage{algorithmic}
\usepackage[algo2e]{algorithm2e}
\usepackage{multirow}
\usepackage{booktabs}
\usepackage{setspace}
\usepackage{caption}
\usepackage{array}
\usepackage{xcolor}
\usepackage{graphicx}
\usepackage{colortbl}

\def\paperID{14153} 
\def\confName{CVPR}
\def\confYear{2025}

\title{SP3D: Boosting Sparsely-Supervised 3D Object Detection via Accurate Cross-Modal Semantic Prompts}

\author{Shijia Zhao$^{1,2}$\footnotemark[1] \quad Qiming Xia$^{1,2}$\footnotemark[1] \quad Xusheng Guo$^{1,2}$ \quad Pufan Zou$^{1,2}$ \quad Maoji Zheng$^{1,2}$ \\ \quad Hai Wu$^{1,2}$ \quad Chenglu Wen$^{1,2}$\footnotemark[2] \quad Cheng Wang$^{1,2}$\\
$^1$Fujian Key Laboratory of Sensing and Computing for Smart Cities, Xiamen University, Xiamen, China\\
$^2$ Key Laboratory of Multimedia Trusted Perception and Efficient Computing,\\
Ministry of Education of China, Xiamen University, China
}

\begin{document}
\maketitle

\renewcommand{\thefootnote}{\fnsymbol{footnote}}
\footnotetext[1]{Equal contribution.}
\footnotetext[2]{Corresponding author.}

\begin{abstract}
Recently, sparsely-supervised 3D object detection has gained great attention, achieving performance close to fully-supervised 3D detectors while requiring only a few annotated instances.
Nevertheless, these methods suffer challenges when accurate labels are extremely absent.
In this paper, we propose a boosting strategy, termed \textbf{SP3D}, explicitly utilizing the cross-modal semantic prompts generated from Large Multimodal Models (LMMs) to boost the 3D detector with robust feature discrimination capability under sparse annotation settings. 
Specifically, we first develop a Confident Points Semantic Transfer (\textbf{CPST}) module that generates accurate cross-modal semantic prompts through boundary-constrained center cluster selection.
Based on these accurate semantic prompts, which we treat as seed points, we introduce a Dynamic Cluster Pseudo-label Generation (\textbf{DCPG}) module to yield pseudo-supervision signals from the geometry shape of multi-scale neighbor points.
Additionally, we design a Distribution Shape score (\textbf{DS score}) that chooses high-quality supervision signals for the initial training of the 3D detector.
Experiments on the KITTI dataset and Waymo Open Dataset (WOD) have validated that SP3D can enhance the performance of sparsely supervised detectors by a large margin under meager labeling conditions.
Moreover, we verified SP3D in the zero-shot setting, where its performance exceeded that of the state-of-the-art methods.
The code is available at \url{https://github.com/xmuqimingxia/SP3D}.
\end{abstract}    
\section{Introduction}
\label{sec:intro}

3D object detection, aiming at locating and classifying objects within 3D scenes, has garnered significant attention in autonomous driving \cite{virconv, voxel-rcnn, bevfusion-mit, 3D-hanet, srkd}. However, the mainstream 3D detectors relies heavily on high-quality labels annotated by humans, which is not only time-consuming but also sensitive to the subjective impression of annotators.

\begin{figure}[t]
  \centering
   \includegraphics[width=1\linewidth]{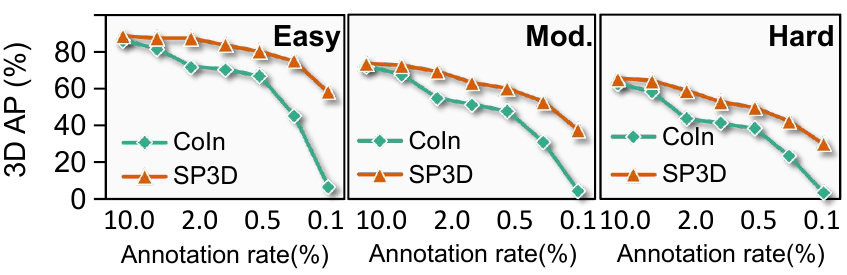}
   \caption{Performance comparison of the sparsely-supervised detector at various annotation rates. SP3D indicates the CoIn initialized with the proposed SP3D.}
   \label{fig:coin_compare}
\end{figure}

To minimize the dependence of 3D detectors on high-quality manual annotations, recent work has begun to focus on label-efficient training strategies \cite{hssda, 3dioumatch, ws3d, coin}.
To discover unlabeled instances, SS3D \cite{ss3d} employs a self-training approach to iteratively optimize the detector trained on sparsely annotated data. 
CoIn \cite{coin} introduces contrastive learning methods, enhancing the model's discriminative capability for various category features. 
However, existing strategies make 3D detectors struggle to extract sufficiently discriminative features from extremely limited annotations. Fig.~\ref{fig:coin_compare} shows some examples where the state-of-the-art (SoTA) sparsely-supervised object detector CoIn \cite{coin} hardly maintains robust performance with a significant reduction in annotation rate.

With the successive emergence and widespread application of large language models (LLMs), such as BERT \cite{bert} and GPT \cite{gpt-3, gpt-4}, in natural language processing, research on large multimodal models (LMMs) is also gaining momentum \cite{clip, glip, sam, groundingdino}. 
Benefiting from the outstanding performance of LMMs, the utilization of pre-trained LMMs has led to significant advancements in 2D vision tasks. Inspired by this, \cite{ulip, pointclip, pointclipv2} transfers the image-text knowledge prior from 2D LMMs to 3D point clouds. However, these methods typically focus on classifying individual instances, and there will be certain limitations when applying them directly to outdoor 3D object detection. Despite this, these attempts to transfer the priors from 2D LMMs to 3D point clouds still provide a new perspective for solving the problem of sparsely-supervised 3D object detection.

\begin{figure}[t]
  \centering
   \includegraphics[width=0.9\linewidth]{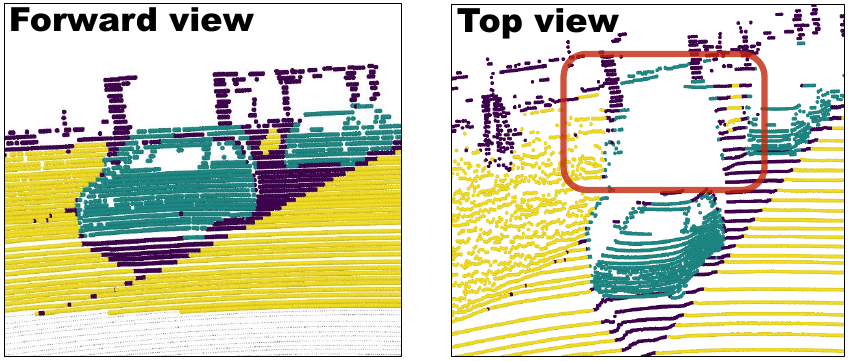}
  
   \caption{Semantics belonging to a foreground object may be incorrectly assigned to background/other objects.
}
   \label{fig:confusion}
\end{figure}

Motivated by these methods, we proposed a two-stage training strategy, \textbf{SP3D}, boosting sparsely-supervised 3D object detection with cross-modal semantic prompts generated by LMMs. 
As shown in Fig.~\ref{fig:overview}, we first employ LMMs to extract semantics from 2D images and explicitly transfer them to 3D point clouds, generating pseudo-labels for the first stage of detector training.
In the second stage, we fine-tune the trained detector with sparse accurate labels. 
However, this enhanced paradigm still holds some challenges: (1) Semantic ambiguity at instance edges. Due to the absence of inherent depth information in images, directly transferring image semantics onto point clouds may result in noisy semantic prompts at the edge of the instance. (2) Efficient generation of pseudo-labels. Based on the obtained semantics, directly fitting pseudo-labels may result in incomplete foreground bounding boxes. Moreover, it is difficult to assess the quality of the generated pseudo-labels because of the lack of ground truth.


To address the aforementioned challenges, our SP3D first designs a Confident Points Semantic Transfer (\textbf{CPST}) module, employing boundary-constrained center cluster selection operation to filter out inaccurate edges to obtain accurate cross-modal semantic prompts. These semantic prompts focus on central foreground semantic masks generated by LMMs. 
Inspired by unsupervised  algorithms~\cite{modest, cpd}, we can utilize these as accurate semantic prompts as seed points to generate bounding box pseudo-labels.
In this case, we introduce a Dynamic Cluster Pseudo-label Generation (\textbf{DCPG}) module and Distribution Shape score (\textbf{DS score}) to efficiently discover and assess high-quality pseudo-labels with complete foreground information from seed points.
As shown in Fig.~\ref{fig:overview}, we utilize the generated pseudo-labels to train the 3D object detector for the first stage. After training, the 3D detector has learned a certain feature discrimination capability from the 2D images. 
Subsequently, we fine-tune the 3D detector with sparse accurate labels, and in conjunction with current label-efficient methods, it demonstrates relatively high detection capabilities even under extremely low labeling scenarios.

Experiment results on the KITTI and Waymo datasets show that SP3D substantially enhances the performance of SoTA sparsely-supervised 3D object detectors. Moreover, without fine-tuning on labeled data, SP3D has also shown superior performance in a zero-shot setting, indicating the efficiency of the SP3D-initialized detector.
The contributions of this paper can be summarized as follows:
\begin{itemize}
    \item We propose a two-stage boosting strategy, \textbf{SP3D}, utilizing accurate cross-modal prompts to boost the feature discrimination capability of 3D detectors under sparsely-supervised situations.
    \item We propose a Confident Points Semantic Transfer (\textbf{CPST}) module for accurate cross-modal semantic prompt generation and constraint, reducing the noise interference in the semantic transfer process. 
    \item We propose a Dynamic Cluster Pseudo-label Generation (\textbf{DCPG}) module and Distribution Shape score (\textbf{DS score}), which generate and assess high-quality pseudo-labels based on semantic prompts to improve the detect performance.
\end{itemize}
\section{Related work}

\subsection{Label-efficient 3D Object Detection}
Recently, label-efficient 3D object detection methods have begun to be explored in responding to the challenge of extremely low annotation volumes.
Generally, these label-efficient methods can be categorized into semi-supervised \cite{3dioumatch,  sess, detmatch, hssda}, and weakly-supervised \cite{vs3d, ws3d, WS3DPR, vit-wss3d} approaches according to the difference in quantity and supervision form.
In semi-supervised methods, full annotations are provided only for some scenes, while weakly-supervised methods typically adopt weaker annotation forms. For example, VS3D \cite{vs3d} leverages detectors on images to guide unsupervised 3D object detection, and WS3D \cite{ws3d} employs point annotation as a supervision signal. 
To maintain accuracy while reducing the annotations, SS3D \cite{ss3d} introduces the concept of sparse supervision, annotating only one complete 3D object per frame.
CoIn \cite{coin} then adopts a contrastive instance feature mining strategy to extract the feature-level pseudo-labels from a significantly reduced amount of annotated data. 
However, the performance of existing methods remains constrained due to the insufficient feature discriminability of the initial detector, which may affect subsequent training under very few annotations.
This work aims to develop a two-stage strategy, enabling the 3D detector to maintain robust feature representation capabilities despite having lower instance annotation ratio.

\subsection{Large Multimodal Models in 3D}
As the outstanding performance achieved by LMMs in 2D tasks \cite{clip, sam, sora, dit, lvm}, some studies have begun to explore their application in the 3D domain.
Inspired by CLIP \cite{clip}, ULIP \cite{ulip} enhances the 3D understanding capability by transferring knowledge from 2D LMM to 3D encoder through contrastive learning methods. Similar works are \cite{pointclip, pointclipv2}.
In the outdoor scenario, SAM3D \cite{sam3d} employs SAM to segment BEV images of point clouds and fit bounding boxes based on the segmentation masks to obtain detection results.
CLIP2Scene \cite{clip2scene} establishes the connection between point clouds and text by using images as an intermediate modality, enhancing the 3D model's semantic understanding of the scene with the prior knowledge of CLIP.
Unlike previous approaches, our SP3D explicitly transfers semantic masks obtained from LMMs onto point clouds to generate high-quality pseudo-labels for the first-stage training of the 3D detector.

\subsection{Multimodal Representation Learning}
Recently, using multimodal methods from 2D images and 3D point clouds to enhance 3D detectors has gradually gained the community's attention \cite{bevfusion-mit, unitr, virconv, contrastalign, sparsefusion}. 
However, these works mainly focus on investigating the image-point cloud fusion strategy, neglecting the utilization of images to explore label-efficient 3D detection. To reduce the required annotations, some methods transfer image information into point clouds to generate pseudo-labels \cite{mixsup}.
However, semantic ambiguity may occur at the object's edge due to the 2D-3D calibration error.
MixSup \cite{mixsup} proposed a connected components labeling strategy, addressing this issue with the spatial separability property inherent to point clouds. SAL \cite{sal} employs density-based clustering to refine imperfect projection issues. Compared with these methods, our SP3D provides a simple but efficient way to reduce semantic noise caused by projection errors.
\section{Methods}

\begin{figure*}[t]
  \centering
   \includegraphics[width=1\linewidth]{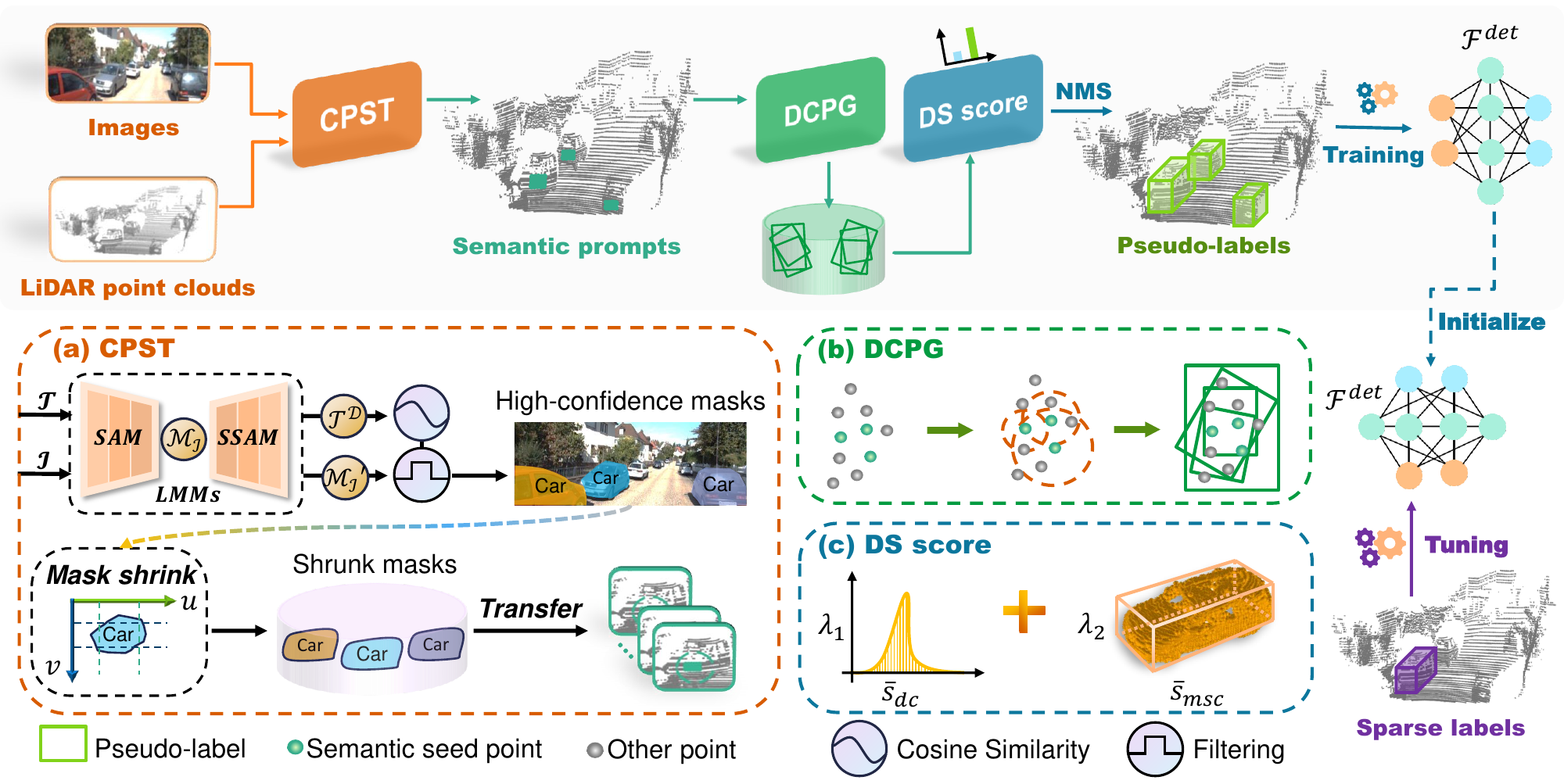}
   \caption{The overview of our SP3D, including (a) CSPT finds semantic seed points through high-confidence semantic masks transfer, (b) DCPG dynamically clusters neighbor points of seed points to fit pseudo-label proposals, and (c) DS score to evaluate the quality of generated pseudo-label proposals, serving as a scoring metric to NMS to suppress low-quality proposals. 
}
   \label{fig:overview}
\end{figure*}

\subsection{Overview}
This paper introduces \textbf{SP3D}, explicitly utilizing the accurate cross-modal semantic prompts generated by LMMs to boost the sparsely-supervised 3D detectors.
As shown in Fig.~\ref{fig:overview}, SP3D consists of three primary components: (1) \textbf{Confident Points Semantic Transfer (CPST)} module, acquiring accurate cross-modal semantic prompts, which we treat as seed points, with the boundary-constrained center cluster selection.
(2) \textbf{Dynamic Cluster Pseudo-label Generation (DCPG)} module, which dynamically generates pseudo-label proposals based on the geometric shapes within the multi-scale neighborhood of these seed points.
(3) \textbf{Distribution Shape score (DS score)}, which employs unsupervised priors as the criterion for evaluating the quality of pseudo-label proposals, and we subsequently apply NMS \cite{nms} to retain high-quality pseudo-labels further.
We then follow the training strategy of CoIn \cite{coin} to train an initial 3D detector with enhanced discriminative capacity. We detail our SP3D framework as follows.

\subsection{Confident Points Semantic Transfer Module}
Encouraged by the development of LMMs, we first utilize the prior knowledge of LMMs to extract semantic information from 2D images explicitly. Meanwhile, by integrating the calibration relationship matrix between 2D images and 3D point clouds, it is straightforward to transfer semantic information onto point clouds. However, as shown in Fig.~\ref{fig:confusion}, there is significant noise in the edge points of instances during the transfer process. 
These semantic points are mistakenly assigned to other instances can have a detrimental impact on subsequent pseudo-label fitting and the training of the detector.
To prevent the incorporation of noise during the transfer of semantic information, we use the boundary-constrained \emph{mask shrink} operation, followed by the coordinate system transformation, to obtain accurate cross-modal semantic prompts.
CSPT is illustrated in Fig.~\ref{fig:overview}(a); we have divided it into two parts as \textbf{LMMs-guided semantic extraction} and \textbf{confident points filtering}.

\paragraph{LMMs-guided semantic extraction.} 
The goal of LMMs is to generate high-quality foreground semantic masks.
Specifically, we take as input an image $\mathcal{I}\in\mathbb{R}^{3\times H\times W}$ and the category-text prompt $\mathcal{T}^{\mathcal{C}}=\left\{t^{\mathcal{C}}_1, t_2^{\mathcal{C}}, ..., t^{\mathcal{C}}_C\right\}$, where $H$ and $W$ denote the height and width of the image, $C$ denotes the categories of interest.
We first utilize FastSAM \cite{fastSAM} to perform  speed-efficient segmentation as 
\begin{equation}
\mathcal{M}_{\mathcal{I}} = \mathrm{SAM}(\mathcal{I}),
\label{eq:sam}
\end{equation}
where $\mathcal{M}_{\mathcal{I}}\in\mathbb{R}^{M\times H\times W}$ denotes the $M$ class-agnostic masks extracted from $\mathcal{I}$, $\mathrm{SAM}(\cdot)$ indicates the FastSAM model. We then utilize $\mathcal{M}_{\mathcal{I}}$ as the mask prompts and feed them, along with image $\mathcal{I}$, into the SemanticSAM \cite{semanticSAM} model, which is expected to output the descriptions $\mathcal{T}^\mathcal{D}=\left\{t^{\mathcal{D}}_1, t_2^{\mathcal{D}}, ..., t^{\mathcal{D}}_M\right\}$ for each mask in $\mathcal{M}_\mathcal{I}$. Specifically:
\begin{equation}
\mathcal{T}^{\mathcal{D}} = \mathrm{SSAM}(\mathcal{I}, \mathcal{M}_{\mathcal{I}}),
\label{eq:ssam}
\end{equation}
where $\mathrm{SSAM}(\cdot)$ refers to the SemanticSAM model. 
Then, we calculate the cosine similarity between the category-text $\mathcal{T}^\mathcal{C}$ and the mask description $\mathcal{T}^{\mathcal{D}}$ to filter out uninteresting background masks, obtaining the foreground masks $\mathcal{M}'_\mathcal{I}$.

\paragraph{Confident points filtering.}
As fuzziness of boundaries results from depth occlusions in images and calibration inaccuracies, we opt to constrain the boundary of foreground masks before 2D-3D transfer, retaining only its central portion.
Specifically, for each foreground mask, we denote its maximum and minimum values in the pixel coordinate system $(u,v)$ as $( u_{min}, u_{max}, v_{min}, v_{max} )$. 
We perform \emph{mask shrink} to constraint boundary range of $\mathcal{M}'_\mathcal{I}$ to obtain $\mathcal{\hat{M}}'_\mathcal{I}$:
\begin{equation}
\begin{split}
& u\in [u_{min}+\frac{1}{2}(1\pm\gamma)(u_{max}-u_{min})],
\\                 
& v\in [v_{min}+\frac{1}{2}(1\pm\gamma)(v_{max}-v_{min})], 
\label{eq:cons}
\end{split}
\end{equation}
where $\gamma$ denotes the shrink factor. With this constraint, we obtain the shrunk masks that filter out the semantically ambiguous regions, ensuring the accuracy of foreground semantic information transferred onto the point clouds. Following \cite{pointpainting}, we transfer the semantic mask from the image onto the point clouds to obtain accurate cross-modal semantic prompts, namely seed points, using the camera's intrinsic and extrinsic parameter matrices. 
Note that, different from \cite{pointpainting}, we adopt the approach of explicitly transferring the shrunk masks onto the point cloud rather than implicitly embedding unprocessed semantic masks into the point cloud's features. This approach helps avoid potential semantic feature confusion between different modalities arising from sparse annotations.

\subsection{Dynamic Cluster Pseudo-label Generation}
With the proposed CPST, we explicitly obtain the semantic seed points from transformation $\mathcal{\hat{M}}'_\mathcal{I}$ of the foreground mask. 
Given $\mathcal{P}\in \mathbb{R}^{N\times 3}=\left\{p_1, p_2, ..., p_N\right\}$ as LiDAR points, we define the seed points covered by $\mathcal{\hat{M}}'_\mathcal{I}$ as $\mathcal{P}_T=\left\{p_t\right\}$, $\mathcal{P}_T \subset \mathcal{P}$. 
It is crucial for the 3D detection task that obtain complete bounding boxes from these seed points.

By referring to traditional unsupervised pseudo-label generation methods \cite{modest}, we produce many pseudo-labels and then use the positional constraints of the seed points to retain the more promising pseudo-labels as supervisory signals. However, existing unsupervised bounding box fitting approaches \cite{modest, cpd} usually take a fixed constant as cluster radius, leading to the problem of inadequate foreground or excessive background noise for the generated bounding boxes.
In this case, we propose a dynamic cluster pseudo-label generation (DCPG) module. This module utilizes the geometry shape of the seed points' multi-scale neighborhood to capture complete foreground information while minimizing background interference. It dynamically generates pseudo-label proposals, as shown in Fig.~\ref{fig:overview}(b).

Specifically, we denote $\mathcal{P}_T^{(k)}$ as the $k$-th instance in a point cloud frame and utilize DCPG dynamically generates a clustering radius $r$ for the $t$-th seed point $p^{(k)}_t$. 
We define the updating function for the dynamic radius as 
\begin{equation}
\mathrm{update}(t, r_{init}) = r_{init}\cdot \frac{t}{N^{(k)}} + \delta, t = 1, 2, ..., N^{(k)},
\label{eq:dcpg}
\end{equation}
where $r_{init}$ is a hyper-parameter set based on empirical experience, $\delta$ denotes adjustment factor to avoid $r$ too small, and $N^{(k)}$ is the number of seed points in current instance.
\begin{algorithm}[H]
    \setlength{\algorithmicindent}{0.5em}
    \caption{Dynamic cluster pseudo-label generation} 
    \label{algo1} 

    \begin{algorithmic}[1]
        \REQUIRE LiDAR points $\mathcal{P}$,  the seed points set $\mathcal{P}_T^{(k)}$ of the $k$-th instance ,  initial radius $r_{init}$, ground remove function $\mathcal{G}$, box fitting function $\mathcal{F}$;
        \ENSURE Pseudo-label proposal set $\mathcal{\hat{B}}^{(k)}$;
        
        \STATE Acquire the number $N^{(k)}$ of points for the seed points set of current instance $\mathcal{P}_T^{(k)}$;
        \STATE $\mathcal{P}_{gr} \leftarrow \mathrm{\mathcal{G}(\mathcal{P})}$ 
        \tcp*{remove gound points}
        \STATE $\mathcal{\hat{B}}^{(k)} = \emptyset$;
        
        \FOR {$t = 1, 2, ..., N^{(k)}$}
            \STATE $p_t = \mathcal{P}_T^{(k)}[t]$ \tcp*{loop each seed point}
            \STATE $r \leftarrow \mathrm{update}(t,r_{init})$
            \STATE $\mathbf{\hat{b}} \leftarrow \mathcal{F}(\mathrm{DBSCAN}(\mathcal{P}_{gr}, p_t, r))$
            \STATE $\mathcal{\hat{B}}^{(k)} \leftarrow \mathcal{\hat{B}}^{(k)}\cup\mathbf{\hat{b}}$
        \ENDFOR 

    \end{algorithmic} 
\end{algorithm}
By applying Eq.~\ref{eq:dcpg}, we dynamically update the radius $r$, $r \in (\delta, r_{init}+\delta]$, during point clustering, thereby obtaining foreground clusters with multi-scale receptive fields.
Following \cite{modest}. We utilize the radius calculated from Eq.~\ref{eq:dcpg} as the clustering radius for DBSCAN \cite{dbscan} and employ \cite{box_fitting} to fit the bounding box for each foreground cluster.
Algorithm~\ref{algo1} summarizes Our DCPG.

\subsection{Distribution Shape Score} 
While DCPG has the capacity for high-quality pseudo-label generation, it unavoidably produces an amount of low-quality pseudo-label proposals. The shapes of these proposals and the extent of foreground completeness contained within the proposals usually deviate significantly from reality. 
Traditional detection methods typically compute the Intersection over Union (IoU) score between predicted bounding boxes and ground-truth (GT) boxes and then employ the NMS \cite{nms} to suppress these low-quality proposals. 
However, lacking GT makes it challenging to directly apply NMS using IoU as the evaluation criterion within our SP3D framework.
In this case, we propose a distribution shape score (DS score) to assess the quality of the pseudo-labels using unsupervised prior knowledge. As illustrated in Fig.~\ref{fig:overview}(c), we divided the DS score into two parts: \textbf{distribution constraint score} and \textbf{meta-shape constraint score}.
 
\paragraph{Distribution constraint score.}
Inspired by \cite{reward}, within a high-quality pseudo-label proposal $\hat{b}$, the distances from its interior points $p_{i, i = 1,...,n}$ to its boundary roughly follow a Gaussian distribution $\mathcal{N}(\mu, \sigma)$, where $\mu = 0.8$ and $\sigma = 0.2$, respectively. 
In other words, we denote random variable $ D = \left \{ d_1,...,d_n \right \}  $ as the distance between $p_i$ and the box boundary of $\hat{b}$, and  $ D \sim \mathcal{N}(0.8, 0.2)$.
Based on this prior, we assign a distribution constraint score to the pseudo-label proposal $\hat{b}$ by calculating the similarity between the random variable $D$ corresponding to each $\hat{b}$ and the normal distribution $\mathcal{N}$.
Specifically: 
\begin{equation}
s_{dc}(\hat{b}) = \frac{1}{|\mathcal{P}_{fg}|} \sum_{p_i\in \mathcal{P}_{fg}} \log( \mathcal{N}(D|\mu, \sigma)),
\label{eq:dcs}
\end{equation}
where $\log(\cdot)$ denotes the logarithm function, $\mathcal{P}_{fg}$ is the foreground points within $\hat{b}$, and $|\mathcal{P}_{fg}|$ represents the number of points in $\mathcal{P}_{fg}$.

\paragraph{Meta-shape constraint score.}
In addition, the shape of a high-quality pseudo-label is expected to be consistent with its template in the real world, which we define as the meta instance, corresponding to its category \cite{cpd}. For class $c$, we denote $\mathcal{B}_c\in\left\{l_c, w_c, h_c\right\}$ as the shape of its meta instance, where $l_c$, $w_c$ and $h_c$ are the normalized length, width and height, respectively.
we followed this shape prior to constructing the class-aware meta-shape constraint score $s_{msc}(\hat{b})$ as
\begin{equation}
s_{msc}(\hat{b}) = 1 - \Phi_{KL}(\mathcal{B}_c|| \hat{\mathcal{B}}_{\hat{b}}),
\label{eq:msc}
\end{equation}
where $\Phi_{KL}(\cdot)$ denotes the normalized KL divergence function, 
and $\hat{\mathcal{B}}_{\hat{b}}\in\left\{l_{\hat{b}}, w_{\hat{b}}, h_{\hat{b}}\right\}$ indicates the normalized shape of the pseudo-label proposal. The purpose of this operation is to suppress the low-quality proposals whose shape deviates significantly from the meta instance.
By combining the distribution constraint score and the meta-shape constraint score, we can obtain the DS score as
\begin{equation}
\mathrm{DS}(\hat{b}) = \lambda_1 \overline{s}_{dc}(\hat{b}) + \lambda_2 \overline{s}_{msc}(\hat{b}),
\label{eq:dc}
\end{equation}
where $\lambda_1$ and $\lambda_2$ denote weight adjustment factor. 
Notably, to unify the dimension, we normalized the two constraint scores before combining them, resulting in $\overline{s}_{dc}$ and $\overline{s}_{msc}$.
We then employ the DS score as a substitution for the confidence score in NMS to suppress the low-quality pseudo-labels.
We utilize the obtained pseudo-labels in conjunction with CoIn \cite{coin} for the first phase of training. Subsequently, we fine-tune the trained detector with a small amount of accurate labels to boost the performance of the 3D detector.
\section{Experiments}
\setlength{\tabcolsep}{6pt}

\begin{table*}[htbp]

\centering
\begin{tabular}{cccccccccccc}
\toprule
\multicolumn{1}{c}{\multirow{2}{*}{Setting}}            & \multicolumn{1}{c}{\multirow{2}{*}{Cost}} & \multicolumn{1}{c}{\multirow{2}{*}{Method}} & \multicolumn{3}{c}{Car}                                                                                     & \multicolumn{3}{c}{Ped}                                                                                     & \multicolumn{3}{c}{Cyc}                                                                 \\ 
\cline{4-12} 
\multicolumn{1}{c}{}                                    & \multicolumn{1}{c}{}                      & \multicolumn{1}{c}{}                        & \multicolumn{1}{c}{Easy}          & \multicolumn{1}{c}{Mod.}          & \multicolumn{1}{c }{Hard}          & \multicolumn{1}{c }{Easy}          & \multicolumn{1}{c }{Mod.}          & \multicolumn{1}{c }{Hard}          & \multicolumn{1}{c }{Easy}          & \multicolumn{1}{c }{Mod.}          & Hard          \\ 
\midrule

\multicolumn{1}{c}{\color[HTML]{6200C9} \textit{Fully-sup.} }                   & \multicolumn{1}{c}{\color[HTML]{6200C9} \textit{100\%}}                 & \multicolumn{1}{c}{\color[HTML]{6200C9} \textit{VoxelRCNN}}               & \multicolumn{1}{c}{\color[HTML]{6200C9} \textit{92.3}}          & \multicolumn{1}{c}{\color[HTML]{6200C9} \textit{84.9}}          & \multicolumn{1}{c}{\color[HTML]{6200C9} \textit{82.6}}          & \multicolumn{1}{c}{\color[HTML]{6200C9} \textit{69.6}}          & \multicolumn{1}{c}{\color[HTML]{6200C9} \textit{63.0}}          & \multicolumn{1}{c}{\color[HTML]{6200C9} \textit{58.6}}          & \multicolumn{1}{c}{\color[HTML]{6200C9} \textit{88.7}}          & \multicolumn{1}{c}{\color[HTML]{6200C9} \textit{72.5}}          & \color[HTML]{6200C9} \textit{68.2}          \\ 
\midrule
\multicolumn{1}{c}{\multirow{5}{*}{Sparsely-sup.}} & \multicolumn{1}{c}{20\%}                   & \multicolumn{1}{c}{SS3D}                   & \multicolumn{1}{c}{89.3}          & \multicolumn{1}{c}{84.2}          & \multicolumn{1}{c}{78.2}          & \multicolumn{1}{c}{-}             & \multicolumn{1}{c}{-}             & \multicolumn{1}{c}{-}             & \multicolumn{1}{c}{-}             & \multicolumn{1}{c}{-}             & -             \\ 
\multicolumn{1}{c}{}                                    & \multicolumn{1}{c}{2\%}  & \multicolumn{1}{c}{VoxelRCNN}                    & \multicolumn{1}{c}{70.5}          & \multicolumn{1}{c}{54.9}          & \multicolumn{1}{c}{44.8}          & \multicolumn{1}{c}{42.6}          & \multicolumn{1}{c}{38.5}          & \multicolumn{1}{c}{32.1}          & \multicolumn{1}{c}{73.3}          & \multicolumn{1}{c}{47.8}          & 43.2          \\  
\multicolumn{1}{c}{} 
& \multicolumn{1}{c}{2\%}  & \multicolumn{1}{c}{CoIn}                    & \multicolumn{1}{c}{89.1}          & \multicolumn{1}{c}{70.2}          & \multicolumn{1}{c}{55.6}          & \multicolumn{1}{c}{50.8}          & \multicolumn{1}{c}{45.2}          & \multicolumn{1}{c}{39.6}          & \multicolumn{1}{c}{80.2}          & \multicolumn{1}{c}{52.3}          & 48.6          \\ 
\multicolumn{1}{c}{}                                    & \multicolumn{1}{c}{2\%}                      & \multicolumn{1}{c}{CoIn++}                  & \multicolumn{1}{c}{\textbf{92.0}}          & \multicolumn{1}{c}{79.5}          & \multicolumn{1}{c}{71.5}          & \multicolumn{1}{c}{46.7}          & \multicolumn{1}{c}{36.1}          & \multicolumn{1}{c}{31.2}          & \multicolumn{1}{c}{82.0}          & \multicolumn{1}{c}{58.4}          & 54.6          
\\  

\multicolumn{1}{c}{}                                    & \multicolumn{1}{c}{2\%}                      & {CoIn++ with SP3D}           & 91.3 & \textbf{80.5} & \textbf{74.0} & \textbf{67.4} & \textbf{58.7} & \textbf{50.9} & \textbf{92.5} & \textbf{73.1} & \textbf{68.3} \\

\midrule
\end{tabular}
\caption{
Comparison with SoTA sparsely-supervised methods on KITTI $val$ split. All methods are based on VoxelRCNN, and we report the 3D AP results of full cost (100\%) and limited cost (20\%, 2\%). The best sparsely-supervised methods are highlighted in \textbf{bold}.
}

\label{tab:sota_comp}
\end{table*}
\setlength{\tabcolsep}{6pt}
\begin{table*}[htbp]
\centering
\begin{tabular}{cc|c|cccccc}
\toprule
\multirow{2}{*}{Stage}        & \multirow{2}{*}{Label} & \multirow{2}{*}{Methods}    & \multicolumn{3}{c}{Car-3D @IoU 0.7} & \multicolumn{3}{c}{Car-BEV @IoU 0.7} \\ 
                              &                        &                            & Easy   & Mod     & Hard    & Easy    & Mod     & Hard    \\ 
                              \hline
\multirow{4}{*}{Single-stage} 
                              & Sparse                 & 1. CenterPoint              & 49.69  & 31.55   & 25.91   & 56.78   & 42.50   & 34.14   \\
                              & Sparse                 & 2. CoIn (CenterPoint-based)& 72.03  & 54.82   & 43.77   & 87.20   & 73.54   & 66.03   \\
                              & Sparse                 & 3. 2 with SP3D (CenterPoint-based)   & 87.44  & 69.24   & 58.61   & 92.72   & 80.00   & 69.01   \\
                              & -                      & 4. \textit{Improvements 3$\to $2}        &\cellcolor[HTML]{C0C0C0} +15.41  &\cellcolor[HTML]{C0C0C0} +14.42   &\cellcolor[HTML]{C0C0C0} +14.84   &\cellcolor[HTML]{C0C0C0} +5.52    &\cellcolor[HTML]{C0C0C0} +6.46    &\cellcolor[HTML]{C0C0C0} +2.98    \\ 
                              \hline
\multirow{4}{*}{Two-stage}               
                              & Sparse                 & 1. Voxel-RCNN            & 70.52  & 54.97   & 44.82   & 83.67   & 71.14   & 57.71   \\
                              & Sparse                 & 2. CoIn (Voxel-RCNN-based)  & 84.56  & 68.47   & 58.02   & 92.31   & 81.01   & 70.24   \\
                              & Sparse                 & 3. 2 with SP3D (Voxel-RCNN-based)    & 91.37  & 74.89   & 63.84   & 95.41   & 85.27   & 74.57   \\
                              & -                      & 4. \textit{Improvements 3$\to $2}        &\cellcolor[HTML]{C0C0C0} +6.81   &\cellcolor[HTML]{C0C0C0} +6.42    &\cellcolor[HTML]{C0C0C0} +5.82    &\cellcolor[HTML]{C0C0C0} +3.1     &\cellcolor[HTML]{C0C0C0} +4.26    &\cellcolor[HTML]{C0C0C0} +4.33    \\ 
                              \hline
\multirow{4}{*}{Multi-stage}  
                              & Sparse                 & 1. CasA                     & 74.18  & 57.37   & 45.05   & 85.90   & 73.21   & 57.23   \\
                              & Sparse                 & 2.CoIn (CasA-based)         & 89.17  & 75.32   & 62.98   & 95.99   & 85.02   & 72.47   \\
                              & Sparse                 & 3. 2 with SP3D (CasA-based)          & 91.12  & 75.94   & 66.46   & 96.55   & 85.65   & 76.31   \\
                              & -                      & 4. \textit{Improvements 3$\to $2}        & \cellcolor[HTML]{C0C0C0}+1.95   & \cellcolor[HTML]{C0C0C0}+0.62    & \cellcolor[HTML]{C0C0C0}+3.48    & \cellcolor[HTML]{C0C0C0}+0.56    & \cellcolor[HTML]{C0C0C0}+0.63    & \cellcolor[HTML]{C0C0C0}+3.84    \\ 
                              \bottomrule
\end{tabular}
\caption{
Comparison with different fully-supervised methods. Sparse label refers to the use of \textit{limited} split (2\% annotation cost). The 3D object detection and BEV detection benchmark are evaluated by mAP with R40, under IoU threshold 0.7.
}
\label{tab:ful_comp}
\end{table*}
\begin{table}[ht]
\setlength{\tabcolsep}{6pt}
\centering
\begin{tabular}{c|cc|cc}
\toprule
\multirow{2}{*}{Methods} & \multicolumn{2}{c|}{Veh. LEVEL 1} & \multicolumn{2}{c}{Veh. LEVEL 2} \\
                         & mAP           & mAPH         & mAP          & mAPH         \\ \hline
CoIn                     & 39.6              & 39.0             & 34.2             & 33.6             \\
CoIn+SP3D                 & 46.7          & 46.0         & 40.4         & 39.9         \\
\emph{Improvement}              & +7.1              & +7.0             & +6.2             & +6.3             \\ \bottomrule
\end{tabular}
\caption{
Comparison with SoTA sparsely-supervised methods on WOD validation set. All methods are based on CenterPoint, and we report the results at a 1\% annotation cost and 0.7 IoU threshold.
}

\label{tab:sota_waymo_comp}
\end{table}

\subsection{Dataset and metrics}
\textbf{KITTI dataset.}
The KITTI \cite{kitti} dataset is a widely used benchmark for 3D object detection.
During the first training stage, we relied solely on the semantic information provided by LMMs to generate pseudo-labels instead of the human-annotated ground truth.
In the fine-tuning stage, we split the training set (7,481 scenes) into a \emph{train} split (3,712 scenes) and a \emph{val} split (3,769 scenes). We follow \cite{coin} to
randomly select 10\% of the scenes from the \emph{train} split and retain only one instance annotation per scene.
In this case, we can obtain a \emph{limited} split, which merely takes 2\% of instance annotations compared with the origin \emph{train} split. All results are evaluated using the official evaluation metric: 3D Average Precision (AP) across 40 recall thresholds (R40).

\textbf{Waymo Open Dataset (WOD).}
The WOD is a widely used large-scale benchmark that contains 798, 202, and 150 sequences (approximately 200 frames per sequence) for training, validation, and testing, respectively. 
All results are evaluated by using 3D mean Average Precision (mAP) and its variant weighted by heading accuracy (mAPH).

\setlength{\tabcolsep}{5pt}

\begin{table*}[htbp]

\centering
\begin{tabular}{c|c|clll|llll|llll}
\toprule
\multicolumn{1}{l|}{\multirow{2}{*}{}} & \multirow{2}{*}{Methods} & \multicolumn{4}{c|}{2\%}                     & \multicolumn{4}{c|}{1\%}  & \multicolumn{4}{c}{0.5\%} \\  
\multicolumn{1}{l|}{}                  &                          & \multicolumn{1}{l}{Car} & Ped. & Cyc. & Avg. & Car  & Ped. & Cyc. & Avg. & Car  & Ped. & Cyc. & Avg. \\ \midrule \midrule
\multirow{3}{*}{semi-sup.}                  & HSSDA                    & 81.9                    & 58.2 & 65.8 & 68.6 & 80.9 & 51.9 & 42.3 & 59.5 & 77.8 & 19.5 & 44.1 & 47.2 \\
                                       & HSSDA+SP3D                & 83.2                    & 60.8 & 74.4 & 72.8 & 83.0 & 53.2 & 69.1 & 68.4 & 82.3 & 22.1 & 46.4 & 50.3 \\
                                       & \emph{Improvement} & \cellcolor[HTML]{C0C0C0}+1.3 & \cellcolor[HTML]{C0C0C0}+2.6 & \cellcolor[HTML]{C0C0C0}+8.6 & \cellcolor[HTML]{C0C0C0}+4.2 & \cellcolor[HTML]{C0C0C0}+2.1 & \cellcolor[HTML]{C0C0C0}+1.3 & \cellcolor[HTML]{C0C0C0}+26.8 & \cellcolor[HTML]{C0C0C0}+8.9 & \cellcolor[HTML]{C0C0C0}+4.5 & \cellcolor[HTML]{C0C0C0}+2.6 & \cellcolor[HTML]{C0C0C0}+2.3 & \cellcolor[HTML]{C0C0C0}+3.1 \\
                                       \hline \midrule
\multirow{6}{*}{sparsely-sup.}             & CoIn                     & 72.3                    & 36.9 & 58.4 & 55.8 & 65.3 & 31.5 & 52.2 & 49.6 & 55.2 & 25.5 & 30.5 & 37.0 \\
                                       & CoIn+SP3D                 & 73.4                    & 44.7 & 62.3 & 60.1 & 71.7 & 36.9 & 59.1 & 55.9 & 63.3 & 36.1 & 43.4 & 47.6 \\ 
                                       & \emph{Improvement} & \cellcolor[HTML]{C0C0C0}
                                       +1.1 & \cellcolor[HTML]{C0C0C0}+7.8 & \cellcolor[HTML]{C0C0C0}+3.9 & \cellcolor[HTML]{C0C0C0}+4.3 & \cellcolor[HTML]{C0C0C0}+6.4 & \cellcolor[HTML]{C0C0C0}+5.4 & \cellcolor[HTML]{C0C0C0}+6.9 & \cellcolor[HTML]{C0C0C0}+6.3 & \cellcolor[HTML]{C0C0C0}+8.1 & \cellcolor[HTML]{C0C0C0}+10.6 & \cellcolor[HTML]{C0C0C0}+12.5 & \cellcolor[HTML]{C0C0C0}+10.6 \\
                                       \cline{2-14}
                                       & HINTED                   & 85.1                    & 60.0 & 81.3 & 75.4 & 82.8 & 49.9 & 74.1 & 68.9 & 80.2 & 45.8 & 64.5 & 63.5 \\
                                       & HINTED+SP3D               & 85.5                    & 60.1 & 81.4 & 75.6 & 84.0 & 57.2 & 81.0 & 74.0 & 83.2 & 49.2 & 73.9 & 68.7 \\
                                       & \emph{Improvement} & \cellcolor[HTML]{C0C0C0}+0.4 & \cellcolor[HTML]{C0C0C0}+0.1 & \cellcolor[HTML]{C0C0C0}+0.1 & \cellcolor[HTML]{C0C0C0}+0.2 & \cellcolor[HTML]{C0C0C0}+1.2 & \cellcolor[HTML]{C0C0C0}+7.3 & \cellcolor[HTML]{C0C0C0}+6.9 & \cellcolor[HTML]{C0C0C0}+5.1 & \cellcolor[HTML]{C0C0C0}+3.0 & \cellcolor[HTML]{C0C0C0}+3.4 & \cellcolor[HTML]{C0C0C0}+9.4 & \cellcolor[HTML]{C0C0C0}+5.2 \\
                                       \bottomrule
\end{tabular}
\caption{
Improvement of SP3D in SoTA semi-supervised and sparsely-supervised methods on KITTI $val$ split  under limited cost setting.  The results are evaluated by mAP with R40, under IoU thresholds 0.7, 0.5, and 0.5 for 'Car', 'Pedestrian', and 'Cyclist', respectively. Additionally, the AP is computed across the difficulty levels of easy, moderate, and hard.
}

\label{tab:diff_sup_diff_ratio}
\end{table*}
\begin{table}[ht]
\setlength{\tabcolsep}{0.3mm}
\centering

\begin{tabular}{c|cc|cccc}
\toprule
\multirow{2}{*}{Methods} & \multicolumn{2}{c|}{Modality} &
\multicolumn{3}{c}{Car-3D @IOU 0.5/0.7}  \\ & Train & Test
                        & Easy        & Mod.         & Hard        \\ 
                        \hline
VS3D  & L+C & L                 & 40.32/9.09  & 37.36/5.73  & 31.09/5.03  \\
WS3DPR & L+C & L              & -/60.01     & -/44.48     & -/36.93     \\ \textbf{SP3D (Ours)} & L+C & L & \textbf{93.75}/\textbf{69.71} & \textbf{76.36}/\textbf{48.65} & \textbf{71.01}/\textbf{40.53} \\ \bottomrule
       \end{tabular}
       \caption{Comparison of zero-shot methods on KITTI $val$ split.}
        \label{tab:zero_kitti}
\end{table}

\subsection{Implementation details}
\paragraph{Pseudo-labels generation.} We directly employed the model parameters provided by FastSAM \cite{fastSAM} and SemanticSAM \cite{semanticSAM} for inference without additional supervisory signals for fine-tuning.
To achieve accurate segmentation results, we set a higher segmentation threshold of $0.7$ during the FastSAM inference process. To mitigate computational demands, we opted to generate pseudo-labels within a confined spatial domain of the semantically relevant points, specifically within an 8-meter radius. We set mask shrink factor $\gamma$ to $0.3$, initial cluster radius $r_{init}$ to $1$, adjustment factor $\delta$ to $0.1$.
We utilize unsupervised priors to filter out the low-quality pseudo-labels that are evidently inconsistent with the intuitive expectations and set the weight adjustment factor of DS score $\lambda_1$ and $\lambda_2$ as $0.5$, $0.5$.

\paragraph{Detector training.} 
We conduct all experiments with a batch size of 8 and a learning rate of 0.003 on 4 RTX 3090 GPUs.
Following previous sparsely-supervised 3D object detection methods \cite{coin, ss3d}, we choose three different classical detectors \cite{centerpoint, voxel-rcnn, casA} as our architecture. And we employ the OpenPCDet \cite{openpcdet} to conduct our experiments. In the first training stage, we employ CoIn \cite{coin} to train an initial detector with the generated pseudo-labels. Then, we use \textit{limited} split to fine-tune the detector.

\paragraph{Baselines.}
To thoroughly validate the effectiveness of SP3D, we select the SoTA sparsely-supervised method \cite{coin} as the primary baseline for comparison. We compare the proposed SP3D approach with the baseline under conventional sparse settings with 2\% annotation cost.
We also compared with cross-modal weakly-supervised methods ~\cite{vs3d, WS3DPR} in a zero-shot setting, which also incorporate visual models to extract semantic information to enhance the performance of detectors.
Furthermore, we establish baselines under progressively reduced annotation costs to evaluate the sensitivity to annotation costs.
\begin{table}[ht]
\setlength{\tabcolsep}{6pt}
\centering

\begin{tabular}{c|cc|ccc} 
\toprule
\multirow{2}{*}{Methods} & \multicolumn{2}{c|}{Modality} & \multirow{2}{*}{Veh.} & \multirow{2}{*}{Ped.} & \multirow{2}{*}{Cyc.} \\
& Train & Test \\
\hline 
SAM3D & L & L & 19.1 & 0.0 & 0.0  \\
CM3D  & L + C & L & 23.7 & 0.1 & 2.6 \\
\textbf{SP3D (Ours)} & L + C & L & \textbf{35.0} & \textbf{15.7} & \textbf{5.9}  \\

\bottomrule
\end{tabular}
\caption{Comparison with zero-shot methods on the WOD validation set. We report Level-2 mAP with IoU thresholds of 0.7, 0.5, and 0.5 for Vehicle, Pedestrian, and Cyclist, respectively.}
\label{tab:zero_waymo}
\end{table}

\subsection{Comparison with SoTA Methods}
\paragraph{Comparison with sparsely-supervised methods.} We integrate our proposed SP3D into the SoTA sparsely-supervised 3D detection algorithm, CoIn++~\cite{coin}. For a fair comparison, all detectors employ the VoxelRCNN~\cite{voxel-rcnn} as the base architecture. As illustrated in Tab.~\ref{tab:sota_comp}, SP3D significantly improves the detection performance of CoIn++. 
The experimental results on WOD, as reported in Tab.~\ref{tab:sota_waymo_comp}, also validate the performance enhancement of our method for the sparsely-supervised 3D object detector.
However, we observe a slight decrease in precision for the 'Easy' car category with our SP3D-initialized model in Tab.~\ref{tab:sota_comp}. This could arise because our initial pseudo-labels are inferred based on the geometric shape of the objects, which may differ from the conventions of manual annotation. When the point cloud structure of an instance is relatively intact, such discrepancies can lead to noticeable differences in the size of the annotated bounding boxes.

\paragraph{Comparison with fully-supervised methods.}
For a fair comparison, we follow CoIn \cite{coin} to select CenterPoint~\cite{centerpoint}, VoxelRCNN~\cite{voxel-rcnn}, and CasA~\cite{casA} as our baseline detectors, representing three distinct types of detection algorithms. 
We initialize the 3D detector using our SP3D, followed by fine-tuning with the \textit{limited} split.
As shown in Tab.~\ref{tab:ful_comp}, detectors designed for full supervision struggle to yield optimal results due to scarce annotations. Despite the effectiveness of CoIn in improving this situation, the results achieved are still unsatisfactory for single-stage detection algorithms with relatively simple structures. 
\begin{table}[ht]
\setlength{\tabcolsep}{8pt}
\centering

\begin{tabular}{c|ccc}
\toprule
\multirow{2}{*}{Methods} & \multicolumn{3}{c}{Car-3D @IoU 0.5} \\
                         & Easy       & Mod.       & Hard      \\ \hline
CPD                      & 61.48      & 61.58      & 60.09     \\
CPD+SP3D                  & 73.65      & 66.13      & 63.78     \\
\emph{Improvement}             & +12.17      & +4.55       & +2.83      \\ \bottomrule
\end{tabular}
\caption{Improvement for unsupervised method.}
\label{tab:cpd_comp}
\end{table}
Our designed strategy, SP3D, significantly reduces this discrepancy, enabling detectors to achieve similar results.

\paragraph{Comparison with zero-shot methods.}
We also compared SP3D with SoTA zero-shot methods. SAM3D \cite{sam3d} and CM3D \cite{cm3d} generate 3D pseudo-labels by leveraging large models for the semantic segmentation of 2D images. In line with the zero-shot approach of prior methods, no 3D annotations were used for training; instead, pseudo-labels were generated from large models and image data. As shown in Tab.~\ref{tab:zero_kitti} and Tab.~\ref{tab:zero_waymo}, the results on the KITTI and Waymo datasets demonstrate that our method significantly outperforms previous approaches.

\paragraph{Comparison with different annotation rates.}
To explore the robustness of SP3D, we conducted experiments on detectors across different supervision paradigms under varying annotation rates, and we reported the results on Tab.~\ref{tab:diff_sup_diff_ratio}. For a fair comparison, we initialize the detectors by following the strategies of CoIn (CenterPoint) \cite{coin}, HSSDA (PV-RCNN) \cite{hssda}, and HINTED (Voxel-RCNN) \cite{hinted}.
The experimental results show that when the annotation rate is limited, detector performance drops sharply as the amount of annotated data decreases. 
By inheriting SP3D, these detectors' performance has been significantly enhanced, mitigating the performance drop associated with reduced annotation.
Taking CoIn as an example, its average AP for the three categories at annotation rates of 1\% and 0.5\% has improved by 6.3\% and 10.6\%, respectively.
These results indicate that even at low annotation rates, SP3D-initialized models can significantly improve the performance of 3D detectors across different supervision paradigms.


\paragraph{Comparison with unsupervised methods.}
To further validate the robustness of SP3D, we integrated it with the SoTA unsupervised detection method CPD \cite{cpd}. As shown in Tab.~\ref{tab:cpd_comp}, despite suffering some impact, the combination of SP3D and CPD still enhances the performance of the unsupervised 3D object detector to a certain extent.

\subsection{Ablation Study}
\paragraph{Effectiveness of mask shrink, DCPG, and DS Score.}
\begin{table}[ht]
\setlength{\tabcolsep}{4pt}
\centering

\begin{tabular}{ccc|ccc}
\toprule
\multirow{2}{*}{Mask shrink} & \multirow{2}{*}{DCPG} & \multirow{2}{*}{DS score} & \multicolumn{3}{c}{Car-3D AP@IoU 0.7}                              \\
                             &                       &                           & Easy                 & Mod.                 & Hard                 \\ \hline
\checkmark &  &  & 35.10 & 23.75 & 19.52  \\
\checkmark & \checkmark &  & 40.56 & 28.15 & 22.40 \\
\multicolumn{1}{l}{} & \checkmark & \checkmark & \multicolumn{1}{l}{47.23} & \multicolumn{1}{l}{33.40} & \multicolumn{1}{l}{27.13} \\
\checkmark & \checkmark & \checkmark & \multicolumn{1}{l}{52.56} & \multicolumn{1}{l}{38.00} & \multicolumn{1}{l}{31.52} \\ \bottomrule
       \end{tabular}
       \caption{Ablation study on KITTI $val$ split. }
        \label{tab:ablation}
\end{table}
To rapidly verify the effectiveness of the proposed modules, we conducted ablation studies based on CenterPoint \cite{centerpoint} and recorded the results in Tab.~\ref{tab:ablation}.
The results presented in the first and second rows illustrate that the precision of pseudo-labels, as augmented by the multi-scale neighborhood clustering mechanism within DCPG, can substantially amplify the detection capabilities of the 3D detector.
This may be attributed to incorporating more comprehensive foreground information in the generated pseudo-labels, which has enhanced the model's feature discrimination capability.
The comparison between the third and fourth rows demonstrates that the mask shrink operation is necessary for handling semantic noise at the instance edges.
Moreover, the results from the second and fourth rows indicate that using the DS score for filtering out low-quality labels can significantly enhance the precision of the detector. 
When combined, the three modules facilitate optimal information transfer and pseudo-label generation, enabling the 3D detector obtained from the first-stage training with more robust performance, and promoting subsequent fine-tuning with accurate labels.

\section{Discussion and 
Conclusion}
This paper proposes a two-stage boosting strategy, SP3D, that leverages accurate cross-modality semantic prompts to enhance the capabilities of sparsely-supervised 3D detectors.
First, we develop a CSPT module to obtain accurate cross-modal semantic prompts, which are treated as seed points, in point clouds by efficiently transferring high-confidence semantic masks extracted with LMMs. 
Next, we introduce a DCPG module to dynamically generate pseudo-label proposals within the multi-scale neighborhoods of seed points. 
Lastly, we propose a DS score as the criterion for NMS to select the high-quality pseudo-labels integrated with the CoIn training strategy to train the initial detector.
After fine-tuning with sparsely annotated data, SP3D demonstrated superior performance over the original sparsely-supervised 3D object detector on the WOD and KITTI dataset, and it maintained robust performance even as the amount of annotation decreased.
\textbf{Limitations:} One limitation is that the current SP3D exhibits a relatively significant degradation when fine-tuning with annotation rates below 0.1\%, which may result from the noise introduced by the extremely low annotations. Future efforts to explore efficient fine-tuning strategies to solve this problem.

\noindent \textbf {Acknowledgment.} This work was supported by the National Natural Science Foundation of China (No.62171393), and the Fundamental Research Funds for the Central Universities (No.20720220064).

{
    \small
    \bibliographystyle{ieeenat_fullname}
    \bibliography{main}
}


\end{document}


\maketitle


\begin{figure}[h]
  \centering
   \includegraphics[width=1\linewidth]{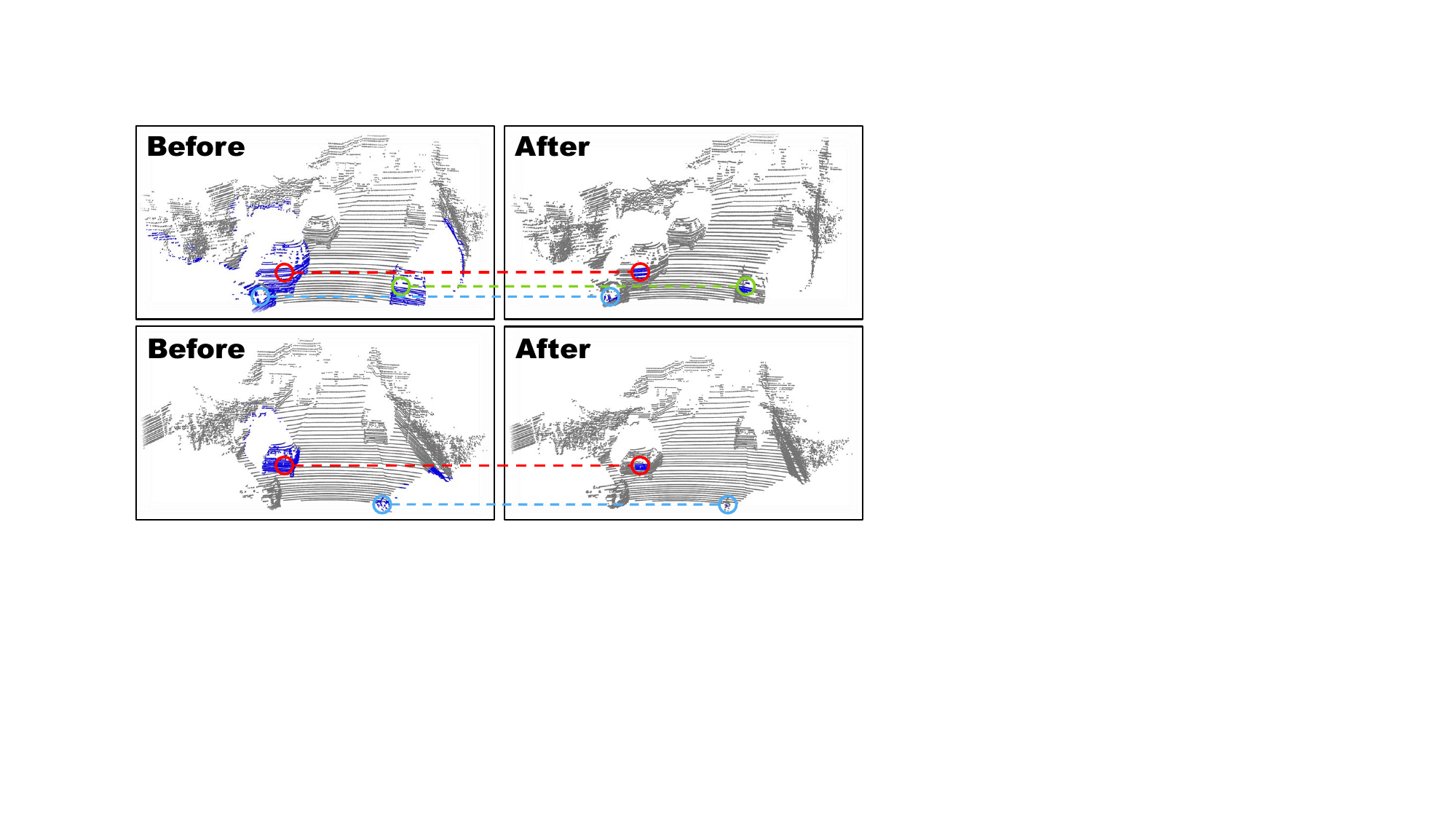}
  
   \caption{Visualization of semantic seed points
   transformed from LMMs-extracted foreground mask. The seed points are colored with \textcolor{blue}{blue}.
}
   \label{fig:mask_shrink}
\end{figure}

\section{The Visualization of Effectiveness of Mask Shrink}
\begin{figure*}[ht]
  \centering
   \includegraphics[width=0.9\linewidth]{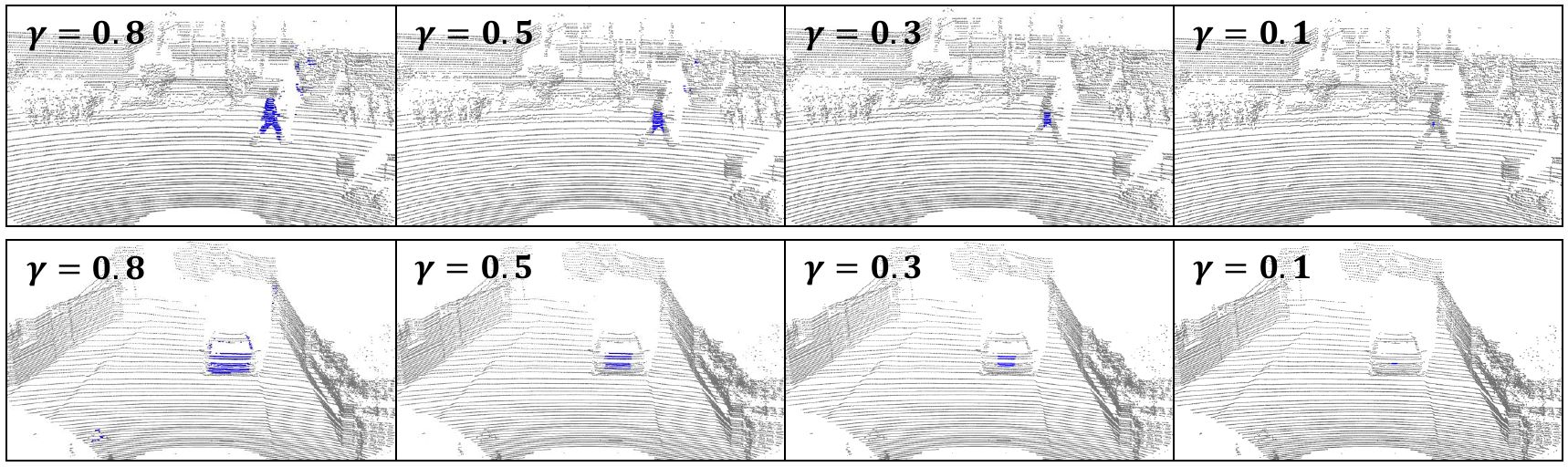}
  
   \caption{The impact of $\gamma$ on mask shrink. The seed points are colored with \textcolor{blue}{blue}.
}
   \label{fig:gamma}
\end{figure*}

Fig.~\ref{fig:mask_shrink} illustrates the impact of the mask shrink operation on the accurate transmission of semantics. For ease of visualization, we have colored the transferred semantic seed points \textcolor{blue}{blue}.
The left column represents directly transferring semantic masks generated by LMMs, where uncertainty edge segmentation, coupled with the inherent one-to-many nature of the pixel-to-point cloud, often results in a significant number of background points being mistakenly classified as foreground.
These pervasive noise exits in seed points significantly hinder the subsequent generation of high-quality pseudo-labels. At the same time, we observe that the noise is primarily concentrated at the edges of the mask. 
Based on this finding, we design a mask shrink strategy based on boundary constraints that only transfer the central region of the foreground masks onto the point cloud, eliminating edge semantic ambiguity and projection uncertainty. After incorporating this module, the effect on the seed points is shown on the right side of Fig.~\ref{fig:mask_shrink}. It can be seen that we finally retained accurate seed points.
\begin{figure}[ht]
  \centering
   \includegraphics[width=1\linewidth]{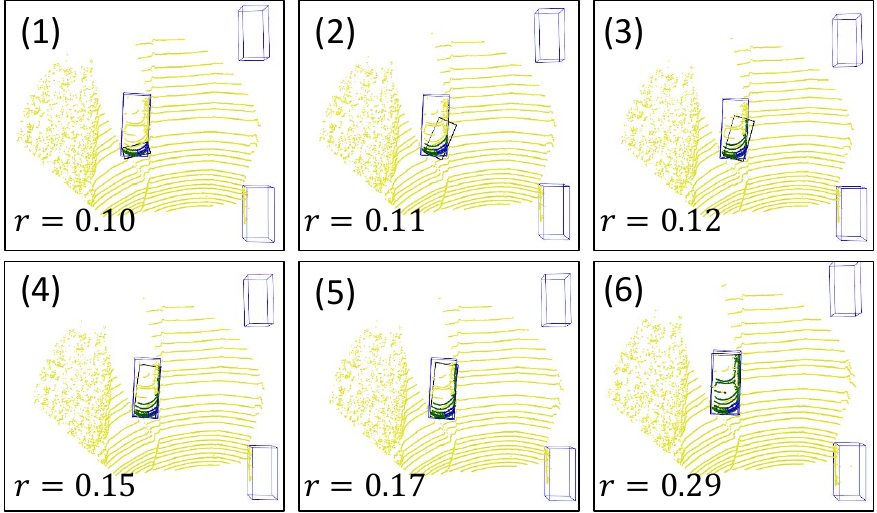}
  
   \caption{Visualization of the process of fitting bounding boxes with dynamic cluster radii in DCPG.
}
   \label{fig:dcpg}
\end{figure}
\paragraph{$\gamma$ on mask shrink.}
The core idea of mask shrink is to filter out the potentially ambiguous edge parts of the foreground mask, retaining only its core area. During the mask shrink process, we set a shrink factor $\gamma$ to control the size of the retained region. We qualitatively analyzed the $\gamma$ values in Fig.~\ref{fig:gamma}.
As shown in the figure, when the $\gamma$ value is too high (e.g. 0.8 and 0.5), it generates a larger number of seed points, but this can lead to significant edge noise. When the $\gamma$ value is too small (e.g. 0.1), the number of seed points is significantly limited, which affects the fitting of subsequent pseudo-labels. Therefore, we have chosen a more balanced value of 0.3 for $\gamma$.

\begin{figure*}[h]
  \centering
   \includegraphics[width=0.9\linewidth]{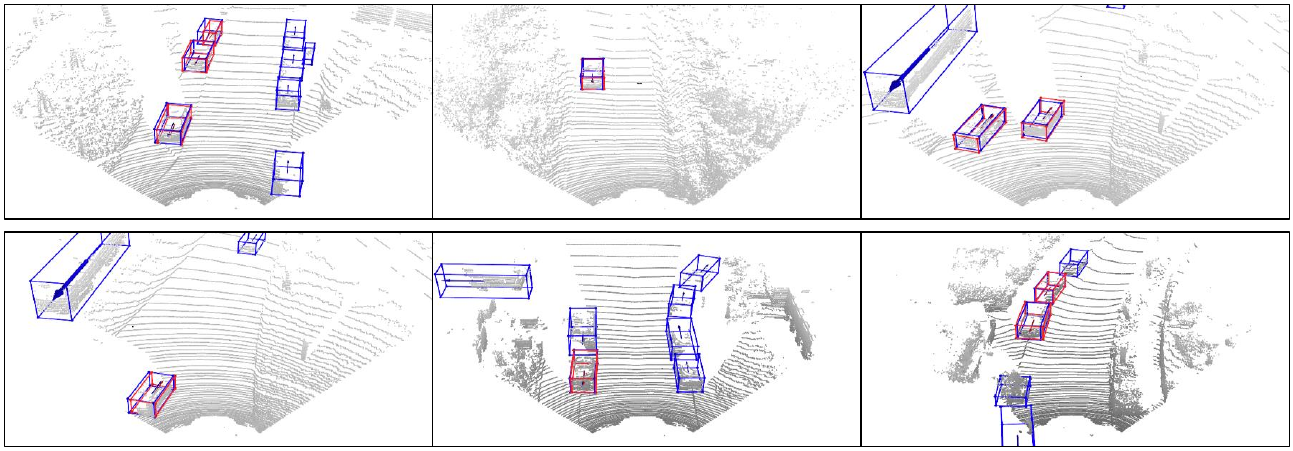}
  
   \caption{Visualization of pseudo-label quality assessment. Bounding boxes with \textcolor{red}{red} color represent the fitted pseudo-labels, while \textcolor{blue}{blue} bounding boxes indicate ground truth boxes.
}
   \label{fig:pseudo_labels}
\end{figure*}
\section{The Visualization of Effectiveness of DCPG}
Fig.~\ref{fig:dcpg} demonstrates the pseudo-label fitting process of DCPG under different clustering radii.
From this example, it can be seen that using a single fixed parameter for the clustering radius $r$ makes it difficult to fit the most appropriate bounding box pseudo-labels. In this case, our DCPG dynamically assigns different cluster radii $r$ to different seed points, which is capable of capturing multi-scale foreground information, thereby fitting higher-quality pseudo-labels.
In addition, by integrating the DS score with the NMS strategy, we can eliminate the low-quality pseudo-labels effectively. Ultimately, it is the high-quality pseudo-labels that remain that provide the necessary support for the training of a high-performing initial 3D detector.

\begin{table}[t]
\setlength{\tabcolsep}{6pt}
\centering
\begin{tabular}{c|ccc}
\toprule
 & $\mathrm{IoU}_{\le 0.5}$ & $IoU_{< 0.7, > 0.5}$ & $\mathrm{IoU}_{\ge 0.7}$\\
\midrule
\midrule
Num. & 156 & 281 & 668 \\
Per. (\%) & 14.12 & 25.43 & 60.45 \\
\bottomrule
\end{tabular}
       \caption{Comparison of pseudo-label quantities across different IoU thresholds.}

\label{tab:recall}
\end{table}
\section{Pseudo-label Quality Assessment}
To visually demonstrate the quality of SP3D-generated pseudo-labels, we simultaneously visualize them with the KITTI GT bounding boxes in Fig.~\ref{fig:pseudo_labels}. 
We represent the pseudo-labels generated by SP3D with the \textcolor{red}{red} boxes and the GT annotations with the \textcolor{blue}{blue} boxes. As shown in the figure, the red boxes exhibit characteristics close to the corresponding blue boxes in the majority of cases, indicating the high quality of the SP3D-generated pseudo-labels.
In addition, to quantitatively analyze the quality of SP3D-generated pseudo-labels, we counted the number of pseudo-labels across various IoU thresholds (0.3, 0.5, 0.7), with the results recorded in Tab.~\ref{tab:recall}. As demonstrated in the table, most of the generated pseudo-labels have an IoU with the GT above 0.7, and the ones with an IoU below 0.5 constitute only 14.12\% of the total, which verifies the effectiveness of our proposed SP3D.

\begin{table}[t]
\setlength{\tabcolsep}{6pt}
\centering
\begin{tabular}{c|ccc}
\toprule
Recall & @IoU 0.3 & @IoU 0.5 & @IoU 0.7 \\ 
\midrule
\midrule
CoIn   & 0.67     & 0.63     & 0.46     \\
+ SP3D    & 0.84     & 0.79     & 0.61     \\ 
\textit{Improvement}    & +0.17     & +0.16     & +0.15 \\
\bottomrule
\end{tabular}
       \caption{The comparison of Recall on different IoU thresholds (0.3, 0.5, 0.7).}

\label{tab:recall_iou}
\end{table}

\begin{table}[ht]
\setlength{\tabcolsep}{7pt}
\centering
\begin{tabular}{cc|ccc}
\toprule
\multirow{2}{*}{Anno. Rate} & \multirow{2}{*}{Method} & \multicolumn{3}{c}{Car-3D @IoU 0.7} \\ 
 & & Easy    & Mod.     & Hard   \\ 
\midrule
\midrule
100\% & CenterPoint & 89.07   & 80.50   & 76.49  \\ 
\midrule
\midrule
\multirow{2}{*}{10\%} & CoIn & 85.95   & 71.80   & 62.64 \\
& + SP3D & 88.84   & 73.56   & 65.17  \\ 
\hline
\multirow{2}{*}{5\%} & CoIn & 81.64   & 67.48   & 58.32  \\
& + SP3D & 87.52   & 72.42   & 63.87  \\ 
\hline
\multirow{2}{*}{2\%} & CoIn & 72.03   & 54.82   & 43.77  \\
& + SP3D & 87.44   & 69.24   & 58.61  \\ 
\hline
\multirow{2}{*}{1\%} & CoIn & 70.39   & 51.31   & 41.31  \\
& + SP3D & 83.79   & 63.16   & 52.50  \\ 
\hline
\multirow{2}{*}{0.5\%} & CoIn & 66.77   & 47.68   & 38.38  \\
& + SP3D & 80.36   & 59.99   & 49.44  \\ 
\hline
\multirow{2}{*}{0.2\%} & CoIn & 45.47   & 31.20   & 23.52  \\
& + SP3D & 75.30   & 52.99   & 42.14  \\ 
\hline
\multirow{2}{*}{0.1\%} & CoIn & 6.84    & 4.65    & 3.61   \\
& + SP3D & 58.57   & 37.41   & 29.88  \\ 
\bottomrule
       \end{tabular}
       \caption{Comparison with different annotation rates (10\% $\to$ 0.1\%). We report the results with 40 recall positions, under
0.7 IoU threshold.}

\label{tab:ratio_kitti}
\end{table}
\section{Comparison with Sparsely-Supervised 3D Detector}
\paragraph{Comparison of recall with different IoU.}
To verify the positive impact of the proposed SP3D on recognition, we evaluated the recall rates under different IoU thresholds. As depicted in Tab.~\ref{tab:recall_iou}, the SP3D model consistently elevates recall rates across the different IoU thresholds, demonstrating a stable improvement. 
Since the geometric information we provide is derived from rule-based generation, a discrepancy exists with the annotated boxes. 
Consequently, this discrepancy results in a slightly higher increase in recall rate at lower IoU thresholds.

\paragraph{Comparison with different annotation rates.}
To more intuitively demonstrate the impact of the proposed SP3D on the sparsely supervised algorithm, we take CoIn \cite{coin} as an example and conduct a group of comparative experiments under different annotation rates. Tab.~\ref{tab:ratio_kitti} provides the variation in performance as annotation rates ranging from 10\% to 0.1\%. 
The experimental results indicate that the original sparsely-supervised 3D detector can significantly enhance performance upon integrating the proposed SP3D. 
For example, at a 2\% labeling rate, the CoIn integrated with SP3D improved 3D AP by 15.41\%, 14.42\%, and 14.84\% on easy, moderate, and hard difficulty levels, respectively. Also, this result represents an average improvement of 14.89\% over the original detector.
Besides, our SP3D significantly boosts the sparsely-supervised 3D detector's performance even at very low annotation rates, which achieves the 41.95\% (36.92\% higher than CoIn) average AP across different difficult levels under the annotation rate of 0.1\%.
The experimental results indicate that the performance of the original sparsely-supervised 3D detector can improve significantly after loading the  SP3D-initialized model, even at low annotation rates.

{
    \small
    \bibliographystyle{ieeenat_fullname}
    \bibliography{main}
}
